\documentclass[sigconf]{acmart}

\usepackage[T1]{fontenc}
\usepackage{lmodern}
\usepackage{booktabs}
\usepackage{multirow}
\usepackage{graphicx}
\usepackage{xcolor}
\usepackage{amsmath}

\usepackage{amssymb}
\usepackage{subcaption}
\usepackage{enumitem}
\usepackage{array}
\usepackage{makecell}
\usepackage{float}
\usepackage{dblfloatfix}
\usepackage{colortbl}
\usepackage{tabularx}
\newcommand{\ie}{\emph{i.e.}}

\newcommand{\etal}{\emph{et al.}}

\newcommand{\raeval}{\textsc{RA-Eval}}
\newcommand{\pp}{\,pp}
\newcommand{\scalarc}{\textsc{scalar-contract}}
\newcommand{\readoutc}{\textsc{extractive-readout}}
\newcommand{\identc}{\textsc{identifier-like}}
\newcommand{\multispanc}{\textsc{compositional-multispan}}
\definecolor{attenbg}{RGB}{255,235,235}
\definecolor{metricoff}{RGB}{40,47,56}
\definecolor{metricjudge}{RGB}{28,39,51}
\definecolor{metricdelta}{RGB}{180,68,62}
\definecolor{metricfn}{RGB}{110,118,128}
\newcolumntype{Y}{>{\raggedright\arraybackslash}X}

\setcopyright{none}
\renewcommand\footnotetextcopyrightpermission[1]{}

\begin{document}

\title[What Does Your Short-Answer VQA Score Actually Measure?]{\texorpdfstring{What Does Your Short-Answer VQA Score Actually Measure?\\Evaluator-Dependent Instability in Multimodal Short-Answer Benchmarks}{What Does Your Short-Answer VQA Score Actually Measure? Evaluator-Dependent Instability in Multimodal Short-Answer Benchmarks}}

\author{
  Guanhua Ye,
  Niu Jingbin,
  Yan Li,
  Meiyu Liang,
  Zhe Xue,
  Yingxia Shao,
  Yawen Li\textsuperscript{*}
}
\affiliation{%
  \institution{Beijing University of Posts and Telecommunications}
  \city{Beijing}
  \country{China}
}
\email{{g.ye, kyungbin, liyanly, meiyu1210, xuezhe, shaoyx, warmly0716}@bupt.edu.cn}
\thanks{*Corresponding author.}

\begin{abstract}
Short-answer VQA benchmarks conflate two distinct quantities: whether a model's answer is semantically correct, and whether that answer matches the surface form expected by the automatic evaluator. We study this conflation across six vision--language models and six benchmarks, using a human-validated semantic judge (97.6\% precision) to audit over 37k official errors. A second text-only judge reproduces the same benchmark-level false-negative pattern, showing that the effect is not an artifact of a single audit model. On text-rich benchmarks, up to half of these errors are semantically acceptable answers penalized purely for surface-form mismatch. This instability is structured by answer type: extractive and multi-span answers are far more evaluator-sensitive than scalar answers. Benign prompt and context rewrites further destabilize official outcomes, flipping item-level correctness at substantial rates without changing the underlying task. A deterministic CPU-only contract repair confirms that the undercount is partially recoverable. These findings imply that official short-answer VQA scores should be accompanied by semantic audits and answer-type diagnostics to remain interpretable.
\end{abstract}



\maketitle
\fancyhead{}
\fancyhead[RO,LE]{\thepage}

\begin{figure}[t]
\centering
\caption{\textbf{Evaluator-sensitive undercount.}}
\label{fig:mm_fig1}
\Description{A two-panel introductory figure. The left panel shows a benchmark question whose canonical answer form is accepted by both the official evaluator and the judge. The right panel shows a semantically acceptable variant answer rejected by the official evaluator but accepted by the judge. The figure also notes that the study covers six benchmarks, forty-one thousand seven hundred twenty-four open-weight questions, and six models.}
\vspace{-0.4em}
\end{figure}

\section{Introduction}\label{sec:intro}

A model looks at a photograph of a jet on a runway and is asked ``What airline is this?'' It outputs ``Air France''---two words, correctly capitalized, unambiguously identifying the carrier. The gold answer is ``Airfrance''---one word, no space. The benchmark marks it wrong. This is not a failure of visual understanding; the model read the livery correctly. It is a failure of the evaluation contract: the answer is semantically acceptable, but its surface form does not survive the string-matching filter. Figure~\ref{fig:mm_fig1} shows this case alongside a second example where a highway-sign question is answered correctly in both form and content.

Short-answer VQA benchmarks have driven much of the progress in text-rich multimodal understanding over the past decade. ST-VQA~\cite{biten2019stvqa} and TextVQA~\cite{singh2019textvqa} stress scene text, DocVQA~\cite{mathew2021docvqa} and InfographicVQA~\cite{mathew2022infographicvqa} extend the challenge to documents and infographics, and ChartQA~\cite{masry2022chartqa} adds visual-logical reasoning over charts. Despite these differences, they share one evaluation habit inherited from the original VQA setup~\cite{antol2015vqa}: compare the emitted string against one or more gold references under an automatic metric. Exact match, ANLS~\cite{peer2024anls_star}, and relaxed numeric accuracy differ in detail, but all make a string-level comparison decisive.

That design fit an era of short, label-like outputs. It fits modern free-form generators less well. Once models answer in natural language rather than from a fixed vocabulary, answer wrapping (``The answer is 42''), morphology, and stylistic variation become part of the measured outcome. Ji~\etal~\cite{ji2024flexeval} argue that rigid exact-match patterns penalize correct but differently worded answers, and Ging~\etal~\cite{ging2024openended_vqa} likewise find substantial disagreement among exact-match, relaxed, and judge-based metrics on open-ended VQA.

The consequence is that a leaderboard number---say, 84.3\% on ST-VQA---is not pure question-answering success. It is the share of outputs that survive a benchmark-specific comparison rule applied to a benchmark-specific reference set. Some rejected outputs are genuine failures, but some are semantically acceptable answers that miss the benchmark's preferred surface form. That gap is the object of this paper.

The answer turns out to depend on what kind of question is being asked. A numeric question like ``How many people?'' has a small set of acceptable answer strings: ``4'', perhaps ``four'', and not much else. A text-reading question like ``What does the sign say?'' has a much larger set: ``STOP'', ``Stop'', ``Stop sign'', ``The sign says STOP'', ``It reads STOP''---all semantically correct, but only some will survive exact match. We call this structural variation the \emph{answer contract}: the implicit agreement between the benchmark and the model about what form an acceptable answer should take. Different answer contracts---scalar, OCR readout, identifier codes, multi-span lists---create different evaluator sensitivities, and benchmarks that mix these contracts in different proportions end up with different amounts of evaluator-dependent undercount.

This matters even when coarse model ranking survives. If official scores undercount semantic success by different amounts across benchmarks and answer types, then adjacent score gaps become harder to interpret, official error pools become contaminated by false failures, and cross-benchmark comparisons stop measuring a common quantity. A model that appears equally strong on two datasets may in fact be passing through two different evaluator filters. Prompt sensitivity compounds the problem: recent work has shown that language models are sensitive to prompt phrasing~\cite{lu2022fantastically,sclar2024quantifying,ye2026rethink,du2026medfuse,zhou2022filter}, and we show that the same sensitivity extends to multimodal short-answer evaluation---benign prompt changes can flip 6--23\% of question-level outcomes without changing the underlying task.

That makes this a measurement problem~\cite{li2013gaussian} rather than a minor reporting nuisance. Benchmark users treat official scores as summaries of model quality, official error pools as summaries of failure, and cross-benchmark deltas as summaries of where a model is strong or weak. If evaluator rejection rates vary sharply across answer contracts, then all three summaries become harder to read: a small score gain can mix semantic improvement with better contract survival, an error pool can contain many acceptable answers, and two benchmarks can assign the same score to very different mixtures of semantic success and surface-form compliance.

We study this problem systematically across six VLMs (four open-weight, two closed-source) and six benchmarks totaling over 41{,}000 questions, organized around four research questions: how prevalent benchmark undercount is after semantic audit (RQ1), how strongly it follows answer contract rather than benchmark name alone (RQ2), how much benign prompt/context changes destabilize official outcomes (RQ3), and whether the resulting mismatch is recoverable with lightweight contract repair (RQ4). To organize these analyses, we introduce an operational answer-contract decomposition (Table~\ref{tab:contract_tree}).

\section{Related Work}\label{sec:related}

\paragraph{VQA evaluation}
Exact-match evaluation was established by Antol~\etal~\cite{antol2015vqa} and remains standard for many extractive VQA benchmarks~\cite{biten2019stvqa,singh2019textvqa,mathew2021docvqa,mathew2022infographicvqa,masry2022chartqa, xu2013image}, even as neighboring tasks have moved toward task-specific relaxations such as ANLS or relaxed numeric accuracy. Recent work has argued for more flexible evaluation of generative VQA responses~\cite{ji2024flexeval} and more fine-grained diagnosis of LVLM evaluation failures~\cite{jiang2024haleval}, while open-ended benchmarks such as MM-Vet already rely on LLM judges for answer-style robustness~\cite{yu2024mm}. Our focus is the short-answer corner where official evaluation is still dominated by exact-match- or ANLS-style benchmark scoring.

\paragraph{Generative answering, ambiguity, and robustness}
Large multimodal generators such as Flamingo, BLIP-2, and LLaVA normalized free-form answering rather than fixed-label prediction~\cite{alayrac2022flamingo,li2023blip2,li2025febt,zhang2026large}. This shift connects short-answer VQA to older tensions between extraction and multiple valid realizations~\cite{rajpurkar2016squad,min-etal-2020-ambigqa,ding2022vdoc,ji2024onetomany}. Prompt sensitivity in LLMs is well documented~\cite{lu2022fantastically,chen2026empowering,zhou2026probdiffflow}, and the same instability extends to VQA evaluation once models are free to verbalize answers~\cite{wang2023selfconsistency,yoran2023answering}.

\paragraph{Evaluation mismatch versus hallucination}
Our question is different from hallucination analysis. Hallucination work asks whether a model invents unsupported visual content~\cite{li2023evaluating,jiang2024haleval}; we ask when an answer may be semantically acceptable yet still be rejected because it fails the benchmark's preferred answer contract. A benchmark error pool can therefore mix genuine task failures, unsupported answers, and semantically acceptable but non-canonical answers. It also interacts with known judge-model biases~\cite{zheng2023lmsys,chen-etal-2024-humans,li-etal-2024-split,koo-etal-2024-benchmarking, guan2021federated}, which is why we treat official metrics as benchmark-facing anchors and use audits only as diagnostic layers.

\section{Framework}\label{sec:framework}

\subsection{Official Score as Semantic Success Filtered by Contract Compatibility}

For question $q$, image $I$, and prompt/context condition $p$, a model produces an output string
\begin{equation}
  o = G_{\theta}(q, I, p),
\end{equation}
and we distinguish two text-level quantities:
\begin{itemize}[leftmargin=1.5em]
  \item \textbf{$S_{\mathrm{sem}}(o)$}: semantic acceptability of the emitted answer text;
  \item \textbf{$\kappa(o)$}: \emph{benchmark-contract compatibility}, \ie{} how well that emitted text matches the answer form preferred by the official evaluator.
\end{itemize}
Conceptually, we summarize the relationship as
\begin{equation}
  S_{\mathrm{off}}(o) = f\!\bigl(S_{\mathrm{sem}}(o), \kappa(o)\bigr),
\end{equation}
where $S_{\mathrm{off}}$ is the benchmark-facing official score and $f$ is a conceptual mapping. This notation separates two quantities that benchmark papers often conflate: whether the emitted text is semantically acceptable at all, and whether that same text survives the benchmark's answer contract.

Model ability governs which answer text $o$ can be produced, but the observed official score is a benchmark-facing measurement after that text passes through contract compatibility.

The value of this notation is diagnostic rather than metaphysical. It does not claim that compatibility is the only thing that matters, nor that every VQA item has one natural contract. It gives us a compact way to ask where benchmark-facing rejection comes from: from answers that are semantically wrong, from answers that are semantically acceptable but contract-incompatible, or from mixtures of both. The empirical sections use this split to connect benchmark-level score deltas to concrete answer regimes rather than treating all official errors as the same kind of failure.

\subsection{Operational Answer Contracts}\label{sec:regimes}

To explain why benchmark sensitivity varies so strongly across datasets, we introduce a coarse operational decomposition by \emph{answer contract}. The split is rule-based and intentionally pragmatic: it uses question prefix, answer shape, and raw benchmark metadata (OCR spans for TextVQA; answer type for InfographicVQA) to group items into a few large buckets that are statistically stable.

\begin{table}[t]
\centering
\caption{\textbf{Operational four-bucket answer-contract taxonomy.} The split is intentionally coarse and heuristic; its purpose is to organize evaluator-sensitive answer regimes into a few reviewer-auditable buckets rather than to define a universal ontology of VQA answers. Residual open-form answers are handled separately in the text and appendix rather than promoted to a fifth named bucket.}
\label{tab:contract_tree}
\small
\setlength{\tabcolsep}{4pt}
\renewcommand{\arraystretch}{1.08}
\begin{tabularx}{\columnwidth}{@{}>{\raggedright\arraybackslash}p{0.24\columnwidth} Y@{}}
\toprule
\textbf{Bucket} & \textbf{Typical answers} \\
\midrule
scalar & numbers, dates, yes/no \\
readout & OCR spans, names, written text \\
identifier & codes, room IDs, alphanumeric strings \\
multi-span & lists, conjunctive answers, multi-span lookups \\
\bottomrule
\end{tabularx}
\end{table}

\begin{itemize}[leftmargin=1.5em]
  \item \textbf{\scalarc{}}: numeric, boolean, or otherwise tightly closed answers;
  \item \textbf{\readoutc{}}: OCR-style span extraction and explicit readout questions;
  \item \textbf{\identc{}}: code-like, uppercase, or short alphanumeric answers;
  \item \textbf{\multispanc{}}: composed answers that aggregate multiple spans or list items.
\end{itemize}

We treat these as \emph{explanatory buckets}, not as a claim that all VQA items fall into four natural kinds. Compatibility typically breaks through numeric wrappers in \scalarc{}, phrasal wrapping in \readoutc{}, separator and code-shape brittleness in \identc{}, and delimiter or composition drift in \multispanc{}. Residual open-form answers remain important, but we treat them as a remainder rather than a fifth symmetric bucket because they are less internally coherent than the four named contracts.

\paragraph{Construction rules}
The rules are intentionally simple and applied in priority order using question form, answer shape, and benchmark-native metadata. Table~\ref{tab:contract_tree} reports the four named buckets used for headline claims, while \textsc{other-open} is treated as an explanatory remainder; finer subtypes remain appendix material except in the contract-repair analysis, where the residual bucket is refined once more for intervention.

\subsection{Measuring Benchmark Undercount}\label{sec:detection}

We use two complementary methods:

\textbf{Heuristic (\raeval{}):} Using canonical multi-answer references when they are available (for example TextVQA's 10 reference answers), an answer is recovered when some gold $g_{\text{norm}}$ appears as a word-boundary substring of $o_{\text{norm}}$, with $|g_{\text{norm}}| > 3$ characters. This is a conservative lower bound on non-canonical but acceptable answers. It is deliberately narrow: it captures cases where answer content survives but answer realization misses the evaluator's preferred contract.

\textbf{LLM-as-Judge audit:} All 36{,}357 error items from open-weight models, plus closed-source model errors on completed datasets, are submitted to DeepSeek-Chat with a binary \emph{text-only} semantic-equivalence prompt. The judge never sees the image; it sees only the question, the reference answer(s), and the model output, which is standard in text-side generative VQA evaluation~\cite{ji2024flexeval,yu2024mm}. We validate this judge with a stratified human review of 570 official-error items spanning all six benchmarks, all six models, and the full audit bucket partition (the four named buckets plus residual \textsc{other-open}). Table~\ref{tab:audit_summary} reports \textbf{97.6\% precision}, \textbf{95.5\% recall}, and \textbf{Cohen's $\kappa = 0.917$} against the human-majority label. Judge precision remains high in every bucket, ranging from 95.1\% in residual \textsc{other-open} to 100\% in \readoutc{} and \identc{}. To verify that benchmark-level undercount patterns are not artifacts of a single judge model, we replicate the full 37{,}033-item audit with a second text-only judge (\texttt{moonshot-v1-8k}). The two judges agree on \textbf{89.5\%} of items ($\kappa = 0.760$), and their per-dataset false-negative rates are almost perfectly aligned (\textbf{$r = 0.999$}), despite different overall acceptance levels (DeepSeek 31.5\%, Moonshot 33.7\%). We use the judge as a semantic-acceptability diagnostic, not as direct evidence of unchanged visual grounding. Together, the heuristic and judge give us a practical decomposition of benchmark score into benchmark-facing acceptance, lower-bound recoverable mismatch, and audited semantic undercount.

\begin{table}[t]
\centering
\caption{\textbf{Judge validation and cross-judge robustness on the official error pool.}}
\label{tab:audit_summary}
\small
\setlength{\tabcolsep}{4pt}
\renewcommand{\arraystretch}{1.05}
\begin{tabular}{@{}lr@{}}
\toprule
\textbf{Metric} & \textbf{Value} \\
\midrule
Precision & 97.6\% \\
Recall & 95.5\% \\
Cohen's $\kappa$ & 0.917 \\
Precision 95\% CI & 95.3--98.8 \\
\midrule
Other-open precision & 95.1\% \\
Scalar precision & 97.6\% \\
Readout precision & 100.0\% \\
Identifier precision & 100.0\% \\
Multi-span precision & 97.1\% \\
\midrule
Agreement & 89.5\% \\
Cross-judge $\kappa$ & 0.760 \\
Per-dataset FN-rate $r$ & 0.999 \\
DeepSeek / Moonshot FN rate & 31.5\% / 33.7\% \\
\bottomrule
\end{tabular}
\end{table}

\section{Experimental Setup}\label{sec:setup}

\begin{figure*}[t]
\centering
\includegraphics[width=\textwidth]{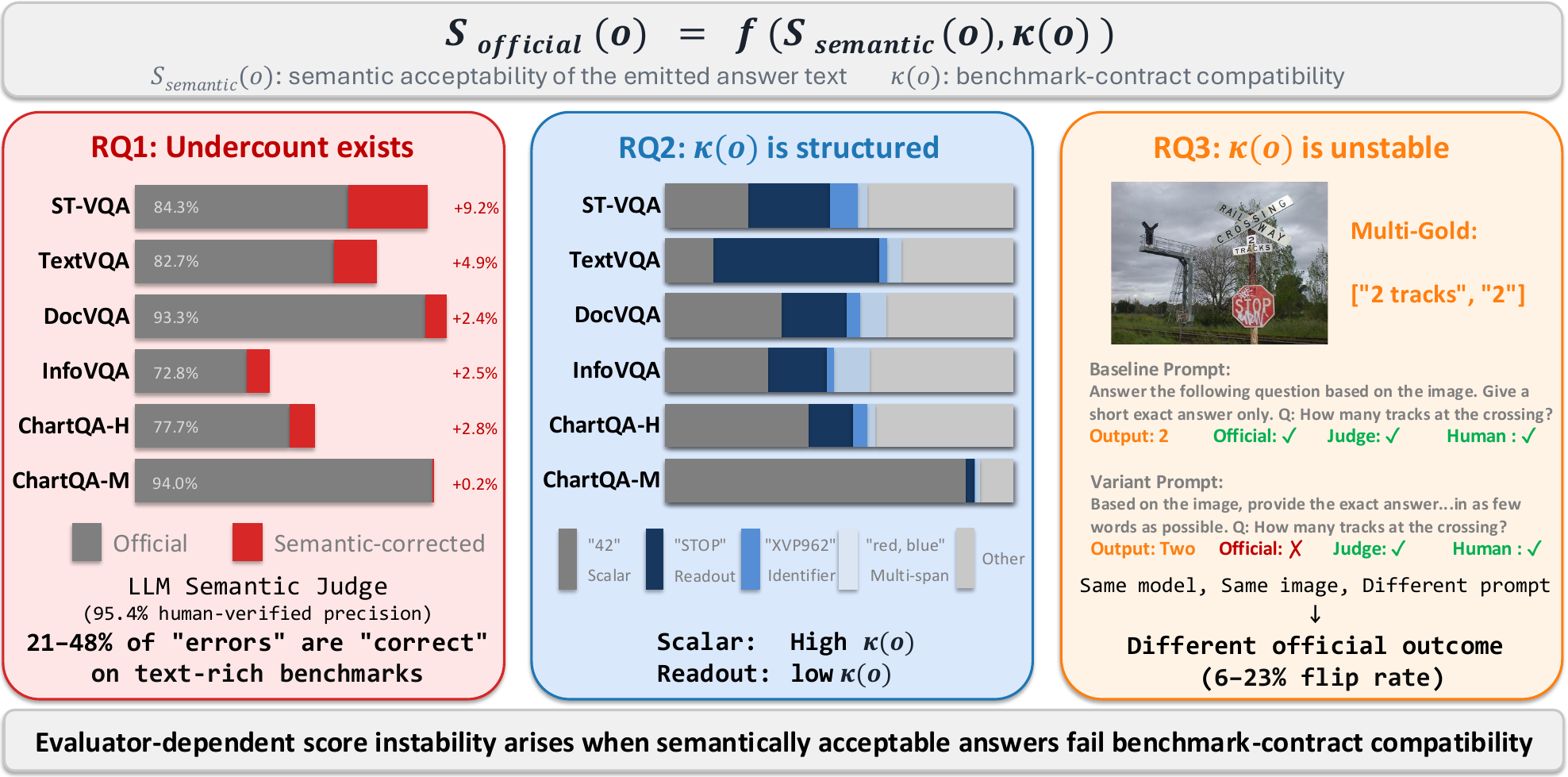}
\caption{\textbf{A summarization of benchmark undercount, contract structure, and perturbation-induced instability.} }
\label{fig:mm_fig3}
\Description{A wide summary figure placed in the experimental setup section. It presents the core conceptual decomposition of official score into semantic acceptability and contract compatibility, then sketches how RQ1 measures benchmark undercount, RQ2 studies answer-contract structure, and RQ3 studies prompt-sensitive instability.}
\end{figure*}

\paragraph{Models}
Four open-weight VLMs: \textbf{InternVL3-8B}~\cite{zhu2025internvl3}, \textbf{Qwen2.5-VL-7B}~\cite{bai2025qwen25vl}, \textbf{Gemma3-12B}~\cite{team2025gemma3}, \textbf{Llama-3.2-11B-Vision}~\cite{grattafiori2024llama3}. Two closed-source VLMs: \textbf{GPT-5.4} (OpenAI) and \textbf{Claude Sonnet~4.6} (Anthropic), evaluated on stratified samples of 500 items for ST-VQA/TextVQA and 200 items for DocVQA, InfoVQA, and ChartQA; Appendix~\ref{app:closed_ci} reports paired-bootstrap confidence intervals for the judge-minus-official deltas. Open-weight inference uses batched vLLM-style serving~\cite{kwon2023vllm}. All models use greedy decoding with the instruction ``Answer the question using a single word or phrase,'' rather than self-consistency or test-time aggregation~\cite{wang2023selfconsistency,yoran2023answering}, so that score changes reflect answer realization under a fixed decoding policy.

\paragraph{Datasets}
We study six benchmarks. \textbf{ST-VQA}~\cite{biten2019stvqa} contributes 26{,}074 items and \textbf{TextVQA}~\cite{singh2019textvqa} contributes 5{,}000. \textbf{DocVQA}~\cite{mathew2021docvqa} contributes 5{,}349 items, and \textbf{InfographicVQA}~\cite{mathew2022infographicvqa} contributes 2{,}801; both use ANLS. \textbf{ChartQA-Human} and \textbf{ChartQA-Machine}~\cite{masry2022chartqa} contribute 1{,}250 items each under 5\% relaxed numeric accuracy. We keep the two ChartQA splits separate because their gold-answer construction differs (human-authored versus machine-generated), which in turn yields different contract mixes and evaluator sensitivities.

\paragraph{Perturbation protocol}
We evaluate seven benign variants: A1--A5 rewrite the short-answer prompt, while B1--B2 change only the surrounding context formatting. We also include one adversarial control (B3), but exclude it from the main flip-rate analysis. For auditability, the main-text wording is summarized here. Base and A1--A5 all share the suffix ``Question: \{question\} Answer:''; only the instruction prefix changes:
\begin{itemize}[leftmargin=1.2em,itemsep=1pt,topsep=2pt]
  \item Base: ``Answer the following question based on the image. Give a short exact answer only.''
  \item A1: ``Look at the image and answer the question below. Provide a brief, exact answer.''
  \item A2: ``Using the provided image, answer the following question with a short exact response.''
  \item A3: ``Based on the image, provide the exact answer below in as few words as possible.''
  \item A4: ``Examine the image and answer the question. Reply with only the short exact answer.''
  \item A5: ``Refer to the image to answer the following. Give only the precise short answer.''
  \item B1: ``\{question\} Answer in one word or number.''
  \item B2: ``\{question\} Give only the final short answer.''
  \item B3: ``\{question\}.''
\end{itemize}

\paragraph{Canonical overlays}
Historical Phase-1 result files were preserved as-is, and any incomplete multi-reference fields were repaired by \emph{adding} canonical overlays rather than overwriting legacy values. The repair adds canonical gold/answer fields plus canonical official-score fields to 17{,}643 rows across seven files; Appendix~\ref{app:canonical} documents the exact schema and impact.

\paragraph{Metric layers}
We intentionally keep three layers separate. The \emph{benchmark-facing automatic score} is the dataset's native or project-standard automatic evaluator. The \emph{canonical overlay score} restores incomplete historical answer fields so that benchmark-native scoring can be re-applied without overwriting legacy artifacts. The \emph{judge-based score} is a semantic-acceptability audit on the error pool only. This separation matters for interpretation: benchmark comparability comes from the first layer, robustness diagnostics from the second, and undercount estimation from the third.

\paragraph{Code and release}
The contract-repair pipeline are released at \url{https://anonymous.4open.science/r/contract-repair-C23F}.

\paragraph{Score legend}
\begin{itemize}[leftmargin=1.2em,label={},itemsep=1pt,topsep=2pt]
  \item \textbf{Official}: benchmark-facing automatic score for leaderboard comparability.
  \item \textbf{True}: canonical-overlay automatic score after restoring incomplete historical answer fields.
  \item \textbf{Judge}: text-only semantic-acceptability audit on the official error pool only.
\end{itemize}

\paragraph{Research questions}
\begin{itemize}[leftmargin=1.2em,label={},itemsep=1pt,topsep=2pt]
  \item \textbf{RQ1. Undercount}: How prevalent and how large is benchmark undercount after semantic audit?
  \item \textbf{RQ2. Structure}: Does instability follow answer contract more than benchmark name?
  \item \textbf{RQ3. Perturbation sensitivity}: How much do benign variants flip official correctness?
  \item \textbf{RQ4. Contract repair}: Can CPU-only contract repair recover undercount?
\end{itemize}

\section{Results}\label{sec:results}

Unless otherwise stated, \emph{Official} refers to the benchmark-facing automatic score, \emph{True} refers to the canonical-overlay automatic score, and \emph{Judge} refers to the text-only semantic audit applied only on the official error pool.

\subsection{Benchmark Undercount After Semantic Audit (RQ1)}\label{sec:rq1}

Under the framework above, this section quantifies the gap between $S_{\mathrm{sem}}(o)$ and $S_{\mathrm{off}}(o)$: how \emph{prevalent} are semantically acceptable benchmark errors, and how \emph{large} is the resulting benchmark undercount once they are audited? Table~\ref{tab:main} gives the full per-model official versus judge-corrected picture. At the per-error level, the undercount is steep and consistent:

\begin{table*}[t]
\centering
\caption{\textbf{Per-model official and judge-corrected benchmark accuracy.} The two rightmost columns summarize mean judge corrections.}
\label{tab:main}
\footnotesize
\setlength{\tabcolsep}{4pt}
\renewcommand{\arraystretch}{1.08}
\begin{tabular*}{\textwidth}{@{\extracolsep{\fill}}llrrrrrrrr@{}}
\toprule
\textbf{Dataset} & \textbf{Metric} & \textbf{InternVL3} & \textbf{Qwen2.5} & \textbf{Gemma3} & \textbf{Llama-3.2} & \textbf{Sonnet$^\dagger$} & \textbf{GPT-5.4$^\dagger$} & \textbf{Open $\bar{\Delta}$} & \textbf{Closed $\bar{\Delta}$} \\
\midrule
\multirow{4}{*}{ST-VQA}
& \textcolor{metricoff}{Official} & 84.3 & 75.8 & 71.6 & 70.2 & 61.4 & 76.4 & \multirow{4}{*}{\textcolor{metricdelta}{\textbf{+9.2}}} & \multirow{4}{*}{\textcolor{metricdelta}{\textbf{+8.7}}} \\

& \textcolor{metricjudge}{Judge}    & 91.8 & 85.6 & 82.4 & 78.7 & 69.4 & 85.8 & \\

& \textcolor{metricdelta}{$\Delta$} & +7.5 & +9.9 & +10.8 & +8.5 & +8.0 & +9.4 & \\

& \textcolor{metricfn}{Acc.\ Err.} & 47.7 & 40.8 & 38.1 & 28.5 & 20.7 & 39.8 & \\
\midrule
\multirow{4}{*}{TextVQA}
& \textcolor{metricoff}{Official} & 82.7 & 84.3 & 76.6 & 84.0 & 82.0 & 82.6 & \multirow{4}{*}{\textcolor{metricdelta}{\textbf{+4.9}}} & \multirow{4}{*}{\textcolor{metricdelta}{\textbf{+5.0}}} \\
& \textcolor{metricjudge}{Judge}    & 87.5 & 89.4 & 82.8 & 87.4 & 86.6 & 88.0 & \\
& \textcolor{metricdelta}{$\Delta$} & +4.9 & +5.1 & +6.1 & +3.4 & +4.6 & +5.4 & \\
& \textcolor{metricfn}{Acc.\ Err.} & 28.1 & 32.2 & 26.2 & 21.2 & 25.6 & 31.0 & \\
\midrule
\multirow{4}{*}{DocVQA}
& \textcolor{metricoff}{Official} & 93.3 & 93.8 & 87.4 & 91.9 & 88.5 & 95.5 & \multirow{4}{*}{\textcolor{metricdelta}{\textbf{+2.4}}} & \multirow{4}{*}{\textcolor{metricdelta}{\textbf{+3.5}}} \\

& \textcolor{metricjudge}{Judge}    & 94.6 & 96.7 & 89.8 & 94.8 & 92.0 & 99.0 & \\

& \textcolor{metricdelta}{$\Delta$} & +1.4 & +2.9 & +2.4 & +2.9 & +3.5 & +3.5 & \\

& \textcolor{metricfn}{Acc.\ Err.} & 20.2 & 46.1 & 18.9 & 36.0 & 30.4 & 77.8 & \\
\midrule
\multirow{4}{*}{InfoVQA}
& \textcolor{metricoff}{Official} & 72.8 & 80.1 & 57.2 & 68.4 & 58.0 & 95.0 & \multirow{4}{*}{\textcolor{metricdelta}{\textbf{+2.5}}} & \multirow{4}{*}{\textcolor{metricdelta}{\textbf{+3.2}}} \\
& \textcolor{metricjudge}{Judge}    & 75.2 & 82.0 & 60.9 & 70.7 & 62.0 & 97.5 & \\
& \textcolor{metricdelta}{$\Delta$} & +2.3 & +1.8 & +3.7 & +2.3 & +4.0 & +2.5 & \\
& \textcolor{metricfn}{Acc.\ Err.} & 8.5 & 9.2 & 8.6 & 7.2 & 9.5 & 50.0 & \\
\midrule
\multirow{4}{*}{ChartQA-H}
& \textcolor{metricoff}{Official} & 77.7 & 78.4 & 63.4 & 55.3 & 90.5 & 93.5 & \multirow{4}{*}{\textcolor{metricdelta}{\textbf{+2.8}}} & \multirow{4}{*}{\textcolor{metricdelta}{\textbf{+1.8}}} \\

& \textcolor{metricjudge}{Judge}    & 79.9 & 81.4 & 66.2 & 58.2 & 92.0 & 95.5 & \\

& \textcolor{metricdelta}{$\Delta$} & +2.2 & +3.0 & +2.8 & +3.0 & +1.5 & +2.0 & \\

& \textcolor{metricfn}{Acc.\ Err.} & 10.0 & 14.1 & 7.6 & 6.6 & 15.8 & 30.8 & \\
\midrule
\multirow{4}{*}{ChartQA-M}
& \textcolor{metricoff}{Official} & 94.0 & 94.5 & 88.2 & 91.4 & 88.5 & 96.5 & \multirow{4}{*}{\textcolor{metricdelta}{\textbf{+0.2}}} & \multirow{4}{*}{\textcolor{metricdelta}{\textbf{+2.2}}} \\
& \textcolor{metricjudge}{Judge}    & 94.2 & 94.5 & 88.7 & 91.5 & 92.0 & 97.5 & \\
& \textcolor{metricdelta}{$\Delta$} & +0.2 & +0.0 & +0.5 & +0.1 & +3.5 & +1.0 & \\
& \textcolor{metricfn}{Acc.\ Err.} & 4.0 & 0.0 & 4.1 & 0.9 & 30.4 & 28.6 & \\
\bottomrule
\end{tabular*}
\par\vspace{2pt}
{\footnotesize\raggedright $^\dagger$ Closed-source entries come from stratified benchmark samples rather than the full open-weight evaluation pools.\par}
{\footnotesize\raggedright \textcolor{metricfn}{Acc.\ Err.} denotes the share of official errors that are judged semantically acceptable by the text-only judge, not an accuracy-style error rate.\par}
\vspace{-0.4em}
\end{table*}

\begin{itemize}[leftmargin=1.5em]
  \item \textbf{ST-VQA:} 28--48\% of errors are judge-accepted, producing +7.5 to +10.8\pp{} correction.
  \item \textbf{TextVQA:} 21--32\%, with +3.4 to +6.1\pp{} correction.
  \item \textbf{DocVQA:} 19--46\%, with +1.4 to +2.9\pp{} correction (partially absorbed by ANLS).
  \item \textbf{ChartQA-Machine:} 0--4\%, with effectively zero correction.
\end{itemize}

\paragraph{Ranking stability versus score meaning}
Judge correction does not produce wholesale leaderboard inversion in our four open-weight model set: ordering changes only on TextVQA and DocVQA, each through a single adjacent swap. But ranking stability is only a weak sanity check, because on ST-VQA, TextVQA, and DocVQA the mean judge correction (+9.2, +4.9, and +2.4\pp{}) exceeds the tightest adjacent-model official gap (1.4, 0.3, and 0.5\pp{}), so adjacent score gaps can lose clear interpretive meaning even when coarse ordering is preserved.

\paragraph{Per-error decomposition}
On ST-VQA, nearly \emph{half} of InternVL3's errors (47.7\%) are semantically acceptable answers rejected by the evaluator. Even on ChartQA-Machine, 0--4\% of errors fall into this category.

That contamination changes how official error pools should be read. A practitioner who mines ST-VQA errors to decide what to fix next is not looking at a pure set of task failures; they are looking at a mixture of genuine misses and evaluator rejections. The larger the judge-accepted slice becomes, the less reliable it is to interpret official errors as direct evidence about what the model did not understand.

\paragraph{Closed-source models}
The same direction holds for the closed-source samples (Table~\ref{tab:main}; paired-bootstrap confidence intervals in Appendix~\ref{app:closed_ci}).

\subsection{Contract Structure of Score Instability (RQ2)}\label{sec:rq2}

\begin{table}[t]
\centering
\caption{\textbf{Contract-level profile of evaluator sensitivity.} Values aggregate the four open-weight models. Residual \textsc{other-open} answers are omitted here and shown separately in Table~\ref{tab:contract_counts}.}
\label{tab:rq2}
\small
\setlength{\tabcolsep}{4pt}
\renewcommand{\arraystretch}{1.08}
\begin{tabular}{@{}lrrr@{}}
\toprule
\textbf{Bucket} & \textbf{Auto gap (pp)} & \textbf{Judge FN (\%)} & \textbf{$\Delta$True (pp)} \\
\midrule
Scalar     & $-1.8$ & 21.1 & $-10.7$ \\
Readout    & $+3.3$ & 36.4 & $-11.9$ \\
Identifier & $+4.0$ & 21.4 & $-9.5$ \\
Multi-span & $+2.0$ & 39.2 & $-11.8$ \\
\bottomrule
\end{tabular}
\end{table}

\begin{table}[t]
\centering
\caption{\textbf{Full dataset-by-contract counts.} Counts are measured on the baseline benchmark pools before perturbation expansion.}
\label{tab:contract_counts}
\small
\resizebox{\columnwidth}{!}{%
\begin{tabular}{@{}l rrrrrr@{}}
\toprule
\textbf{Dataset} & \textbf{Total} & \textbf{Scalar} & \textbf{Readout} & \textbf{Ident.} & \textbf{Multi} & \textbf{Other} \\
\midrule
ST-VQA    & 26{,}074 & 4{,}707 & 6{,}110 & 2{,}022 & 516 & 12{,}719 \\
TextVQA   & 5{,}000  & 1{,}256 & 2{,}372 & 49    & 119 & 1{,}204 \\
DocVQA    & 5{,}349  & 1{,}396 & 796   & 154   & 770 & 2{,}233 \\
InfoVQA   & 2{,}801  & 1{,}337 & 113   & 20    & 331 & 1{,}000 \\
ChartQA-H & 1{,}250  & 945   & 24    & 4     & 38  & 239 \\
ChartQA-M & 1{,}250  & 1{,}080 & 21    & 6     & 1   & 142 \\
\bottomrule
\end{tabular}
\par}
\end{table}

Table~\ref{tab:rq2} shows where the compatibility term $\kappa(o)$ is systematically low. Evaluator sensitivity is not uniform. Readout answers have a sizable official gap (+3.3\pp{}) and high judge false-negative rate (36.4\%), consistent with evaluator brittleness to OCR-style phrasing. Multi-span answers have the highest judge false-negative rate (39.2\%) and nearly the largest perturbation sensitivity, indicating that composed answers amplify benchmark instability. Identifier-like answers form a separate pocket: they show the largest baseline official gap (+4.0\pp{}) but more moderate perturbation sensitivity than readout or multi-span answers.

Scalar answers show the lowest judge false-negative rate, but they are not immune to perturbation sensitivity. Overall, evaluator sensitivity remains contract-specific rather than benchmark-specific alone.

The contract mix also matters at the benchmark level, and Table~\ref{tab:contract_counts} makes that mix explicit. ChartQA-M is overwhelmingly scalar (86.4\%), whereas TextVQA is almost half readout (47.4\%) and ST-VQA combines a large readout slice with a visible identifier pocket. Table~\ref{tab:contract_counts} therefore confirms that benchmark-level correction magnitude tracks contract composition. Appendix~\ref{app:subtypes} shows exploratory subtypes within the identifier-like and multi-span buckets.

Two anchors keep the taxonomy from collapsing into pure post hoc storytelling. TextVQA exposes OCR token structure directly, and InfographicVQA exposes answer-type annotations that distinguish single-span from multi-span answers. The exploratory subtype evidence in Appendix~\ref{app:subtypes} points in the same direction: \identc{} is driven most strongly by alphanumeric and hyphenated codes, while \multispanc{} is driven most strongly by list-like answers. Appendix~\ref{app:group_compare} adds a harder comparison against numeric/text and dataset-only grouping, and Appendix~\ref{app:contracts} collects further examples.

This is why benchmark name alone is too coarse an explanatory unit. Datasets are mixtures of contracts, not atomic evaluator regimes. A benchmark-level correction can be high because it contains many readout and residual-open questions, low because it is mostly scalar, or intermediate because it mixes several contracts with different sensitivities. The contract view therefore does not replace benchmark identity, but it explains why benchmark-level score gaps differ even when the official metric looks superficially similar.

\subsection{Benign Perturbations Destabilize Official Scores (RQ3)}\label{sec:rq3}

Table~\ref{tab:rq3} reports the mean question-level flip rate over the seven benign variants A1--A5 and B1--B2 for two representative open-weight models; the parenthesized value is the corresponding mean official-score delta over the same seven variants. We show Gemma3 and InternVL3 in the top block for readability, while the family summary in the bottom block averages over all four open-weight models. Flip rates of 1.8--29.4\% indicate that a single benchmark score is not a stable measurement of semantic success under fixed task content.

\begin{table}[t]
\centering
\caption{\textbf{Representative benign-perturbation instability patterns.} Top block: mean question-level flip rates across the seven benign variants for two representative open-weight models shown for readability; parentheses show the corresponding mean official-score delta. Bottom block: family-level mean score drops, averaged over all four open-weight models and six datasets.}
\label{tab:rq3}
\small
\setlength{\tabcolsep}{4pt}
\renewcommand{\arraystretch}{1.06}
\begin{tabularx}{\columnwidth}{@{}>{\raggedright\arraybackslash}p{0.28\columnwidth}YY@{}}
\toprule
\textbf{Dataset} & \textbf{Gemma3} & \textbf{InternVL3} \\
\midrule
ST-VQA & \makecell[c]{6.3\%\\($-2.4$\pp{})} & \makecell[c]{2.9\%\\($+0.3$\pp{})} \\
TextVQA & \makecell[c]{6.7\%\\($-2.6$\pp{})} & \makecell[c]{3.7\%\\($+0.6$\pp{})} \\
DocVQA & \makecell[c]{20.4\%\\($-16.8$\pp{})} & \makecell[c]{8.5\%\\($-6.8$\pp{})} \\
InfoVQA & \makecell[c]{29.4\%\\($-14.8$\pp{})} & \makecell[c]{15.7\%\\($-8.8$\pp{})} \\
ChartQA-H & \makecell[c]{7.1\%\\($-1.5$\pp{})} & \makecell[c]{5.4\%\\($-1.5$\pp{})} \\
ChartQA-M & \makecell[c]{3.1\%\\($-1.3$\pp{})} & \makecell[c]{1.8\%\\($-0.5$\pp{})} \\
\midrule
\textbf{Family} & \textbf{Official $\Delta$} & \textbf{True $\Delta$} \\
A (prompt) & $-6.7$\pp{} & $-2.8$\pp{} \\
B (context) & $-4.7$\pp{} & $+0.5$\pp{} \\
\bottomrule
\end{tabularx}
\end{table}

Excluding the Qwen2.5-VL outlier still leaves mean flip rates of 4.1\% on ST-VQA, 5.5\% on TextVQA, 15.3\% on DocVQA, 22.7\% on InfoVQA, and 2.8--8.0\% on the two ChartQA splits; Appendix~\ref{app:qwen} reports the full Qwen profile separately.

The family block also separates two kinds of instability. Prompt rewrites reduce both official and true scores, which is consistent with changes in the emitted answer itself. Context-only edits reduce official score more than true score, indicating that some flips come from benchmark-facing realization drift rather than uniform task failure.

That split matters because it distinguishes instability in model behavior from instability in evaluation behavior. When both official and true scores move together, the perturbation is changing what the model says. When official drops more than true, the perturbation is changing mainly how benchmark-facing the same underlying answer becomes. The second pattern is particularly damaging for interpretation: the task content is effectively unchanged, but the recorded benchmark outcome still flips.

\subsection{CPU-Only Bidirectional Contract Repair (RQ4)}\label{sec:rq4}

We use contract repair as a diagnostic intervention on the comparison layer itself. The intervention stays entirely on CPU and has three variants: \emph{output-only} conservative rewriting of the baseline output, \emph{reference-only} contract-aware canonicalization of accepted reference forms, and \emph{bidirectional} repair that applies both in the same benchmark-facing comparison. A \emph{break} in this section means a row that was officially correct under the original comparison but becomes incorrect after a repair rule is applied, so net gains should always be read together with the number of introduced breaks.

\begin{table*}[t]
\centering
\caption{\textbf{CPU-only contract repair on the open-weight baseline pools.} Values are $\Delta$Official in percentage points relative to the original baseline score. \emph{Output-only} applies conservative output rewriting; \emph{Reference-only} expands accepted reference forms without changing model outputs; \emph{Bidirectional} combines both in the same contract-normalized comparison.}
\label{tab:rq4_finalizer}
\small
\setlength{\tabcolsep}{5pt}
\renewcommand{\arraystretch}{1.06}
\begin{tabular}{@{}lrrrrrrr@{}}
\toprule
\textbf{By dataset} & \textbf{ST-VAQ} & \textbf{TextVQA} & \textbf{DocVQA} & \textbf{InfoVQA} & \textbf{ChartQA-H} & \textbf{ChartQA-M} & \textbf{Overall} \\
\midrule
Output-only    & +0.35 & +0.23 & +0.05 & +0.07 & +0.14 & +0.00 & +0.26 \\
Reference-only & +1.35 & +0.47 & +0.47 & +0.71 & +1.58 & +0.10 & +1.06 \\
Bidirectional  & \textbf{+1.71} & \textbf{+0.75} & \textbf{+0.52} & \textbf{+0.75} & \textbf{+2.14} & \textbf{+0.10} & \textbf{+1.34} \\
\bottomrule
\end{tabular}

\vspace{0.45em}

\begin{tabular}{@{}lrrrrrr@{}}
\toprule
\textbf{By bucket} & \textbf{Scalar} & \textbf{Readout} & \textbf{Ident.} & \textbf{Multi} & \textbf{Other} & \textbf{Overall} \\
\midrule
Output-only    & +0.07 & +0.14 & +0.10 & -0.01 & +0.49 & +0.26 \\
Reference-only & +1.31 & +0.79 & +1.26 & +3.24 & +0.80 & +1.06 \\
Bidirectional  & \textbf{+1.37} & \textbf{+0.94} & \textbf{+1.36} & \textbf{+3.61} & \textbf{+1.30} & \textbf{+1.34} \\
\bottomrule
\end{tabular}
\end{table*}

Table~\ref{tab:rq4_finalizer} shows that the strongest practical intervention is no longer output-only rewriting. Conservative output repair alone remains small (+0.26\pp{}, net +432, 31 breaks). Reference-side canonicalization is much stronger and break-free: it raises official accuracy by +1.06\pp{} overall, recovering 1,767 official errors without changing any model output. Combining the two yields the best result, +1.34\pp{} overall (net +2,237, only 4 breaks), which corresponds to 6.16\% official-error recovery and 17.44\% judge-false-negative recovery on 166,731 open-weight rows.

The efficiency cost stays low. On our 80-core CPU machine, the full bidirectional pass over all 166,731 open-weight rows completes in about 45 seconds with 32 workers, and none of the three variants reruns a VLM.

The gap between output-only and reference-only repair is itself informative: much of the undercount lives on the reference side of the comparison interface rather than in removable output wrappers alone. The dataset block shows gains on all six benchmarks, led by ChartQA-Human (+2.14\pp{}) and ST-VQA (+1.71\pp{}), while the bucket block shows gains on all five routes, led by multi-span (+3.61\pp{}). The remaining 82.56\% of judge-confirmed false negatives indicate the limit of conservative deterministic repair: many require deeper paraphrase handling, alias resolution, or phrase-level restructuring beyond CPU-only string normalizationn~\cite{xiao2022retromae}.

This also explains why earlier prompting experiments plateaued. If much of the mismatch comes from narrow accepted reference forms, then re-answering the image can help only indirectly; it still asks the model to rediscover an accepted string.

Reference-side canonicalization acts closer to the bottleneck by widening the accepted interface without changing the model output. Bidirectional repair works best because some cases are genuinely output-side wrappers, while others are reference-side omissions.

\subsection{Output-Side Case Studies}\label{sec:rq4_cases}

\begin{figure*}[t]
\vspace{-10pt}
\centering
\includegraphics[width=\textwidth]{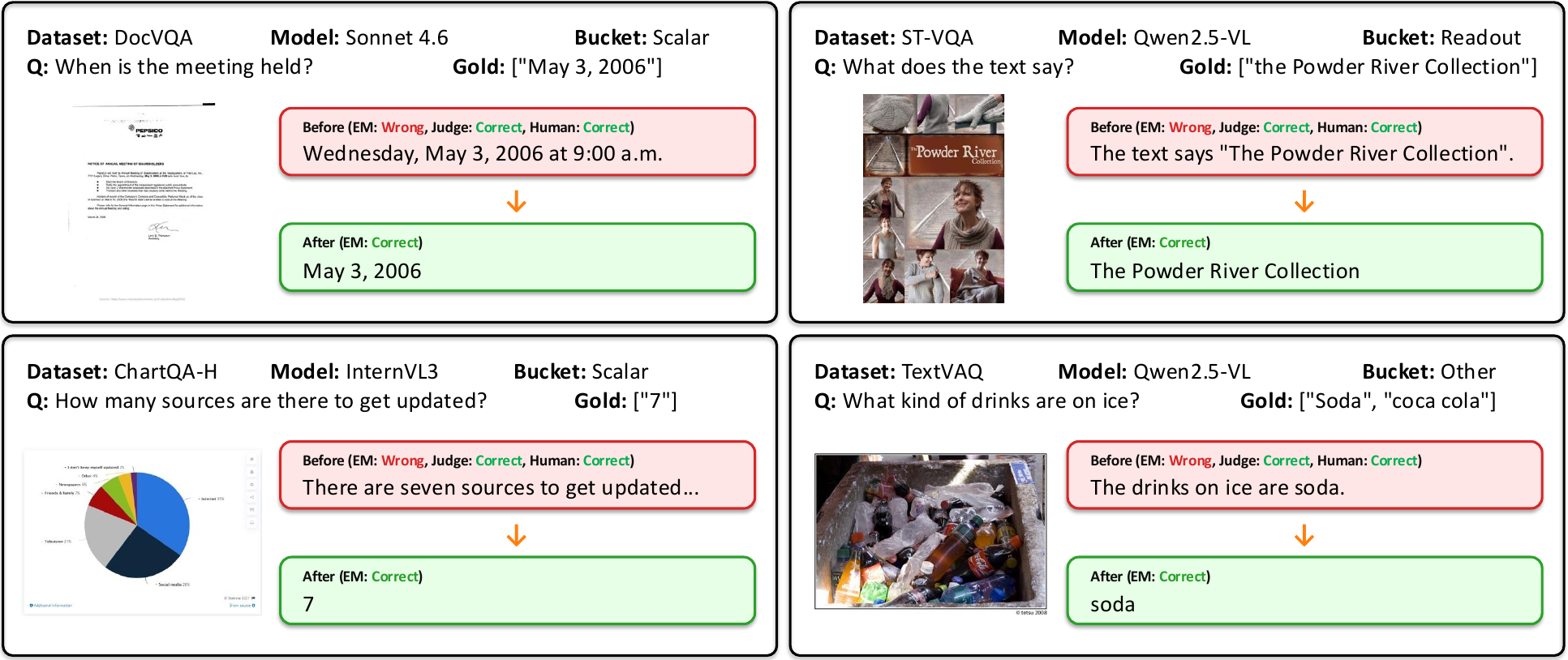}
\caption{\textbf{Representative output-side repairs inside the broader bidirectional pipeline.} All four examples start as official rejects but judge-accepted answers, and all four become official passes after the conservative output-side rewrite.}
\label{fig:rq4_cases}
\vspace{-10pt}
\end{figure*}

Figure~\ref{fig:rq4_cases} shows four representative output-side repairs spanning all four named contracts. In each case, a conservative strip or extraction recovers the benchmark-facing answer form from a judge-accepted but officially rejected output.

\section{Discussion}\label{sec:discussion}

\paragraph{Cross-benchmark comparisons are confounded}
Reported scores are not commensurable across benchmarks with different contract mixes. A model scoring 85\% on ST-VQA and 85\% on ChartQA-M has passed through two different contract-compatibility filters, so a one-point gain cannot be assumed to mean the same thing everywhere. ST-VQA contains a large readout and residual-open component, while ChartQA-M is overwhelmingly scalar. The same reported accuracy can therefore encode very different mixtures of semantic success and contract survival.

\paragraph{What should be reported and designed for}
Official metrics should be retained, but they should not be treated as self-interpreting~\cite{li2019application}. Without semantic audits and contract-aware breakdowns, reported improvements can come from more semantic success, higher contract compatibility, or both. The same logic applies to benchmark design: dataset creators should document not only answer format and scoring rules, but also the \emph{contract mix} of the evaluation pool. For model builders, the practical consequence is equally direct: official error pools should not be mined as if they were clean failure sets. Appendix~\ref{app:reporting_template} gives one compact reporting template implied by our findings.

\paragraph{Limitations}
The LLM-as-Judge audit is strong but not perfect: the 570-item human validation yields 97.6\% precision and 95.5\% recall, with residual disagreements concentrated in residual \textsc{other-open} and scalar cases. Our contract taxonomy is heuristic and operational rather than universal, and some effects remain benchmark-specific. Closed-source evaluation uses 500-item samples on ST-VQA/TextVQA and 200-item samples on the remaining benchmarks, which are sufficient for pattern confirmation but not for fine-grained model ranking. Prompt perturbations do not isolate answer realization from task behavior; our claim is score instability, not preserved latent understanding. More broadly, we study only text-side interface shifts here; extending the contract-aware view to image-side or cross-modal perturbations remains future work.

\section{Conclusion}\label{sec:conclusion}

Short-answer VQA benchmarks do not measure semantic success alone. Across six models and six benchmarks, we show that evaluator-dependent instability is large on text-rich datasets, small on ChartQA-M, and structured by answer contract rather than benchmark name alone. In our current model set, the resulting distortions usually preserve coarse ranking, but they still change the meaning of score gaps, contaminate official error pools, and destabilize item-level outcomes under benign perturbations.

Official short-answer VQA scores are under-interpreted if they are read as pure semantic success. Our CPU-only contract repair further shows that part of this undercount is structurally recoverable. Short-answer VQA papers should keep official metrics, but accompany them with semantic audits and contract-aware diagnostics if they want those scores to remain interpretable.

\bibliographystyle{ACM-Reference-Format}
\bibliography{references}

\clearpage
\nobalance
\appendix

\section{Raw-Split Support}\label{app:contract_support}

Two raw benchmark fields directly support the heuristic split. In TextVQA, 2{,}670/5{,}000 questions have OCR-extractable gold answers, versus 2{,}330 that require non-extractive or canonicalized forms. In InfographicVQA, the raw answer-type metadata contains 1{,}968 \emph{single-span}, 180 \emph{multi-span}, 100 \emph{question span}, and 553 \emph{non-extractive} items. These raw attributes are not sufficient to define the full taxonomy, but they anchor the \readoutc{} and \multispanc{} buckets in benchmark-native metadata rather than only post hoc answer-shape rules.

\section{Why Not Just Numeric versus Text?}\label{app:group_compare}

\noindent\textbf{Small grouping comparison.}
For raw judge-FN separation, a dataset-only grouping is strongest ($\eta^2=0.050$), followed by numeric/text ($0.033$) and the contract taxonomy ($0.024$). For benign perturbation flips, the ordering reverses: the contract taxonomy is strongest ($\eta^2=0.018$), slightly above dataset-only ($0.012$) and numeric/text ($0.011$). This comparison is intentionally modest. We do \emph{not} claim that the contract taxonomy dominates every alternative on every signal. Dataset identity remains the strongest separator of raw judge-FN rates because benchmark-specific annotation and metric choices are real. A numeric/text split also captures a substantial part of baseline evaluator mismatch. The value of the contract view is different: it is the only coarse, cross-benchmark split here that cleanly separates benign perturbation instability while preserving distinct \readoutc{}, \identc{}, and \multispanc{} pockets that the numeric/text split collapses.

\section{Exploratory Contract Subtypes}\label{app:subtypes}

\begin{table*}[t]
\centering
\caption{\textbf{Exploratory subtype evidence within \identc{} and \multispanc{}.} These finer splits are used only as supporting evidence. Main-text claims remain at the four-bucket level.}
\label{tab:subtypes}
\small
\setlength{\tabcolsep}{4pt}
\renewcommand{\arraystretch}{1.05}
\begin{tabularx}{\textwidth}{@{}>{\raggedright\arraybackslash}p{0.15\textwidth} >{\raggedright\arraybackslash}p{0.20\textwidth} >{\centering\arraybackslash}p{0.08\textwidth} >{\centering\arraybackslash}p{0.16\textwidth} >{\centering\arraybackslash}p{0.12\textwidth} >{\centering\arraybackslash}p{0.16\textwidth}@{}}
\toprule
\textbf{Parent} & \textbf{Subtype} & \textbf{$n$} & \makecell[c]{\textbf{Official}\\\textbf{$\rightarrow$ True}} & \textbf{Judge FN\%} & \makecell[c]{\textbf{Perturb}\\\textbf{$\Delta$True}} \\
\midrule
\identc{} & alnum code        & 1{,}480 & +17.1\pp{} & 23.1 & $-10.5$\pp{} \\
\identc{} & compact num+unit  & 708     & +7.4\pp{}  & 22.6 & $-9.9$\pp{} \\
\identc{} & hyphenated code   & 544     & +11.7\pp{} & 27.8 & $-11.2$\pp{} \\
\identc{} & short upper       & 6{,}652 & +0.1\pp{}  & 21.2 & $-9.1$\pp{} \\
\midrule
\multispanc{} & conjunctive   & 976     & +1.2\pp{}  & 36.7 & $-12.3$\pp{} \\
\multispanc{} & list-2        & 1{,}948 & $-1.5$\pp{} & 40.0 & $-10.8$\pp{} \\
\multispanc{} & list-3plus    & 852     & +1.4\pp{}  & 31.9 & $-14.7$\pp{} \\
\multispanc{} & slash pair    & 24      & +27.0\pp{} & 78.6 & $-9.7$\pp{} \\
\bottomrule
\end{tabularx}
\end{table*}

The subtype table clarifies two otherwise hidden patterns. First, the most brittle part of \identc{} is not all short strings, but specifically alphanumeric and hyphenated code-like outputs. Second, the perturbation sensitivity of \multispanc{} is driven primarily by list-like and conjunctive answers rather than by every multi-span surface equally. Because some subtypes are small (notably \emph{slash pair}), we treat these as hypothesis-generating rather than headline claims.

\section{Canonical Overlay Repair}\label{app:canonical}

\begin{table}[t]
\centering
\caption{\textbf{Historical canonical-overlay repair.} Legacy fields were preserved and repaired by adding canonical overlay fields only where raw multi-reference answers were incomplete.}
\label{tab:canonical_overlay}
\small
\resizebox{\columnwidth}{!}{%
\begin{tabular}{@{}lrrr@{}}
\toprule
\textbf{Dataset} & \textbf{Files} & \textbf{Rows} & \textbf{Overlay rows} \\
\midrule
TextVQA  & 6 & 21{,}000  & 5{,}000 \\
ST-VQA   & 6 & 105{,}296 & 8{,}859 \\
DocVQA   & 6 & 21{,}796  & 2{,}390 \\
InfoVQA  & 6 & 11{,}604  & 1{,}394 \\
\midrule
\textbf{Total} & \textbf{24} & \textbf{159{,}696} & \textbf{17{,}643} \\
\bottomrule
\end{tabular}
\par}
\end{table}

This repair removes a misleading TextVQA artifact. After the overlay is applied, all four open-weight models remain \emph{negative} on TextVQA under the canonical true-official score: InternVL3 $-9.13$\pp{}, Qwen2.5-VL $-20.45$\pp{}, Gemma3 $-10.90$\pp{}, and Llama-3.2 $-4.76$\pp{}. The repaired data still supports evaluator-dependent instability, but it no longer supports the earlier TextVQA/Qwen ``official reversal'' story. The main text therefore focuses on benchmark undercount and contract-aware instability rather than stronger claims about invariant latent understanding.

\section{Qwen2.5-VL Prompt Sensitivity}\label{app:qwen}

\begin{table}[H]
\centering
\caption{Qwen2.5-VL accuracy (\%) under perturbation variants. A1/A2 cause catastrophic drops on text benchmarks due to verbose output under minimal instructions.}
\label{tab:qwen_detail}
\small
\begin{tabular}{@{}l rrrrr rr@{}}
\toprule
 & A1 & A2 & A3 & A4 & A5 & B1 & B2 \\
\midrule
ST-VQA   & 40.3 & 65.8 & 76.6 & 77.2 & 76.3 & 70.3 & 79.7 \\
TextVQA  & 25.8 & 61.2 & 83.2 & 86.8 & 85.8 & 83.6 & 89.2 \\
DocVQA   & 39.1 & 76.5 & 86.8 & 87.3 & 87.7 & 83.0 & 88.8 \\
InfoVQA  & 41.9 & 57.4 & 61.7 & 62.0 & 64.2 & 61.4 & 65.1 \\
ChartQA-H & 51.2 & 68.8 & 78.1 & 79.0 & 79.4 & 79.5 & 81.7 \\
ChartQA-M & 67.7 & 75.7 & 94.6 & 94.6 & 93.5 & 94.6 & 95.3 \\
\bottomrule
\end{tabular}
\end{table}

Table~\ref{tab:qwen_detail} clarifies why Qwen2.5-VL is treated separately in the main text. The large drops are concentrated in A1 and A2, where lightly reworded prompts elicit verbose answers on text-rich benchmarks. The later prompt variants and the B-family context changes are much less pathological, so the table is best read as an outlier profile rather than as the central evidence for RQ3.

\section{LLM-as-Judge Precision Audit}\label{app:audit}

The main text reports the headline human-validation statistics: a stratified 570-item review of the official error pool yields 97.6\% precision, 95.5\% recall, and Cohen's $\kappa=0.917$ for the text-only judge against human-majority labels. We keep this appendix only as a bookkeeping note that the reviewed spreadsheet spans all six benchmarks, all six models, and the full audit bucket partition. The central point is unchanged: the judge is sufficiently reliable for a diagnostic role, but it is still treated as a semantic-audit layer on the official error pool rather than as a replacement benchmark metric.

\section{Closed-Source Uncertainty Intervals}\label{app:closed_ci}

\begin{table*}[t]
\centering
\caption{\textbf{Closed-source judge-minus-official deltas with paired-bootstrap 95\% confidence intervals.} These intervals are computed on the sampled closed-source benchmark subsets and therefore quantify sampling uncertainty rather than model stochasticity.}
\label{tab:closed_ci}
\small
\setlength{\tabcolsep}{4pt}
\renewcommand{\arraystretch}{1.05}
\begin{tabular}{@{}lcc@{}}
\toprule
\textbf{Dataset} & \textbf{Claude Sonnet~4.6} & \textbf{GPT-5.4} \\
\midrule
ST-VQA & $+8.0$\pp{} \; [5.8, 10.4] & $+9.4$\pp{} \; [7.0, 12.0] \\
TextVQA & $+4.6$\pp{} \; [2.8, 6.6] & $+5.4$\pp{} \; [3.4, 7.4] \\
DocVQA & $+3.5$\pp{} \; [1.0, 6.0] & $+3.5$\pp{} \; [1.0, 6.0] \\
InfoVQA & $+4.0$\pp{} \; [1.5, 7.0] & $+2.5$\pp{} \; [0.5, 5.0] \\
ChartQA-H & $+1.5$\pp{} \; [0.0, 3.5] & $+2.0$\pp{} \; [0.5, 4.0] \\
ChartQA-M & $+3.5$\pp{} \; [1.0, 6.5] & $+1.0$\pp{} \; [0.0, 2.5] \\
\bottomrule
\end{tabular}
\end{table*}

Table~\ref{tab:closed_ci} shows that the closed-source deltas are directionally consistent with the open-weight results even under finite-sample uncertainty. The widest positive intervals again appear on ST-VQA and TextVQA, while the chart benchmarks remain smaller. The purpose of this appendix table is therefore not to claim a separate closed-source story, but to show that the main undercount pattern is not an artifact of the open-weight pool alone.

\section{Additional Contract Examples}\label{app:contracts}

\begin{table*}[t]
\centering
\caption{Illustrative examples behind the four answer-contract buckets. These examples are not exhaustive; they are included to make the taxonomy auditable.}
\label{tab:contract_examples}
\small
\begin{tabularx}{\textwidth}{@{}>{\raggedright\arraybackslash}p{0.13\textwidth} >{\raggedright\arraybackslash}p{0.26\textwidth} >{\raggedright\arraybackslash}p{0.10\textwidth} Y@{}}
\toprule
\textbf{Contract} & \textbf{Question} & \textbf{Gold} & \textbf{Representative non-canonical output} \\
\midrule
\scalarc{} & When does support for legalization reach the peak? & \texttt{2014} & \texttt{The percentage of people who support legalization reaches the peak in 2014 with 52\%.} \\
\readoutc{} & What is the name of the medication? & \texttt{SEPTRA} & \texttt{SEPTRA Suspension} \\
\identc{} & Where is the women's gym located? & \texttt{C-5} & \texttt{The women's gym is located in C-5.} \\
\multispanc{} & Which two countries are being compared in the graph? & \texttt{[Austria, Chile]} & \texttt{Austria and Chile} \\
\bottomrule
\end{tabularx}
\end{table*}

Table~\ref{tab:contract_examples} makes the four-way split more concrete. The scalar example fails because a correct value is wrapped inside an explanatory sentence; the readout example appends a descriptive suffix to the right OCR token; the identifier-like example embeds a short code inside a full clause; and the multi-span example compresses a list answer into a conjunctive phrase. These are not random formatting accidents. They are the kinds of realization shifts that turn semantically acceptable answers into official benchmark errors, which is why the contract taxonomy is useful as an explanatory device rather than merely a labeling scheme.

\section{Minimal Reporting Template}\label{app:reporting_template}

\begin{table}[H]
\centering
\caption{A minimal reporting template for short-answer VQA. This is the practical protocol implied by our findings.}
\label{tab:reporting_template}
\scriptsize
\renewcommand{\arraystretch}{1.06}
\setlength{\tabcolsep}{2pt}
\resizebox{0.98\columnwidth}{!}{%
\begin{tabular}{@{}>{\raggedright\arraybackslash}p{0.16\columnwidth} >{\raggedright\arraybackslash}p{0.76\columnwidth}@{}}
\toprule
\textbf{Layer} & \textbf{What to report} \\
\midrule
\makecell[l]{Benchmark\\score} & \makecell[l]{Standard dataset metric for comparability\\ with prior work} \\
\makecell[l]{Contract\\mix} & \makecell[l]{Breakdown over scalar, readout, identifier,\\ and multi-span answer buckets} \\
\makecell[l]{Semantic\\audit} & \makecell[l]{Judge or human audit\\ on the official error pool} \\
\makecell[l]{Failure\\notes} & \makecell[l]{Empty outputs, partial answers, and formatting failures,\\ reported separately from semantic mismatch} \\
\bottomrule
\end{tabular}
\par}
\end{table}

Table~\ref{tab:reporting_template} is intentionally minimal. Its role is not to prescribe a new benchmark protocol in full detail, but to make the paper's practical implication concrete: benchmark-facing scores should remain visible, while contract mix, semantic audit, and failure notes provide the diagnostic context needed to interpret those scores responsibly.

\paragraph{Operational taxonomy notes.}
The main text keeps four named buckets plus a residual \textsc{other-open} remainder on purpose. The named buckets have enough support to sustain headline claims, whereas \textsc{other-open} is treated as a heterogeneous remainder and refined into wrapper-open versus semantic-open only inside the contract-repair analysis, not as a replacement of the main taxonomy.

\end{document}